\pdfoutput=1
\documentclass{article}
\usepackage{ijcai19}

\usepackage{times}
\usepackage{soul}
\usepackage{url}
\usepackage[hidelinks]{hyperref}
\usepackage[utf8]{inputenc}
\usepackage[small]{caption}
\usepackage{graphicx}
\usepackage{subfigure}
\usepackage{amsmath}
\usepackage{autobreak}
\usepackage{booktabs}
\usepackage{algorithm}
\usepackage{algorithmic}
\usepackage{hyperref}
\usepackage{cleveref}
\usepackage{bm}
\title{Hierarchical Reinforcement Learning for Multi-agent MOBA Game}

\author{
Zhijian Zhang
\and
Haozheng Li\and
Luo Zhang\and
Tianyin Zheng\and
Ting Zhang\and \\
Xiong Hao\and 
Xiaoxin Chen\and
Min Chen\and
Fangxu Xiao\and
Wei Zhou\\
\affiliations
vivo AI Lab\\
\emails
\{zhijian.zhang, haozheng.li, luo.zhang, zhengtianyin, haoxiong\}@vivo.com
}

\begin{document}

\maketitle

\begin{abstract}
Real Time Strategy (RTS) games require macro strategies as well as micro strategies to obtain satisfactory performance, since it has large state space, action space, and hidden information. This paper presents a novel hierarchical reinforcement learning model for mastering Multiplayer Online Battle Arena (MOBA) games, a sub-genre of RTS games. The contributions are: (1) proposing a hierarchical framework, where agents execute macro strategies by imitation learning and carry out micromanipulations through reinforcement learning, (2) developing a simple self-learning method to get better sample efficiency for training, and (3) designing a dense reward function for multi-agent cooperation in the absence of game engine or Application Programming Interface (API). Finally, various experiments have been performed to validate the superior performance of the proposed method over other state-of-the-art reinforcement learning algorithms. Agents successfully learn to combat and defeat bronze-level built-in AI with 100\% win rate, and experiments show that our method can create a competitive multi-agent for a kind of mobile MOBA game {\it King of Glory} in 5v5 mode.
\end{abstract}

\section{Introduction}

Deep reinforcement learning (DRL) has become a promising tool for game AI since its success in playing game Atari \cite{mnih2015human}, AlphaGo ~\cite{silver2017mastering}, Dota 2 ~\cite{OpenAI}, and so on. Researchers verify algorithms by conducting experiments in games quickly, and transfer this ability to real world applications such as robotics control, recommend services. Unfortunately, there are still many challenges in practice. More and more researchers have started to conquer more complex Real Time Strategy (RTS) games such as StarCraft and Defense of the Ancients (Dota) recently. Dota is a kind of Multiplayer Online Battle Arena (MOBA) game which includes 5v5 and 1v1 modes. To achieve victory in a MOBA game, players need to control their only one agent to destroy enemies’ crystal.

\begin{figure}[tb]
\vskip 0in
\begin{center}
\centering
\subfigure[5v5 map]{
\label{Fig.sub.1}
\includegraphics[width=0.47 \columnwidth]{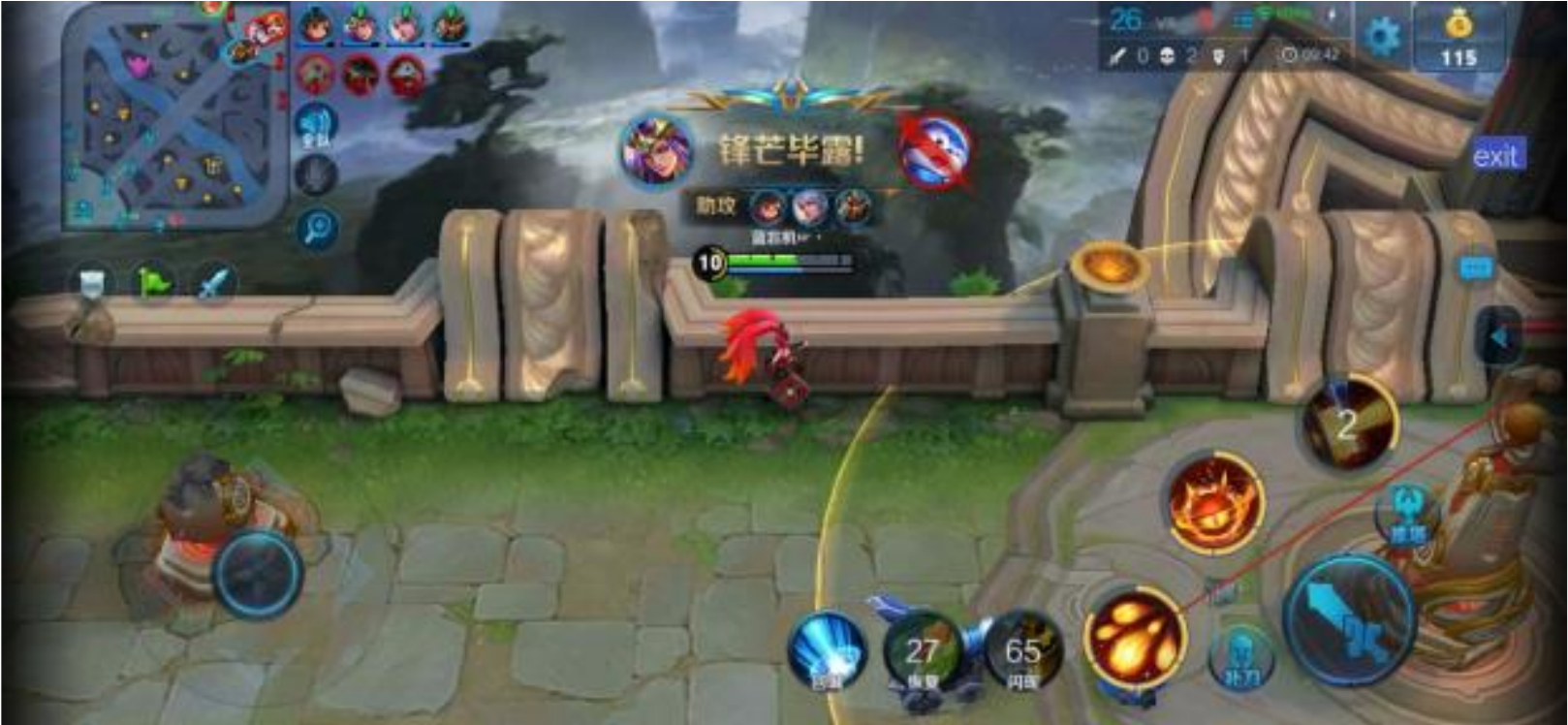}}
\subfigure[1v1 map]{
\label{Fig.sub.2}
\includegraphics[width=0.47 \columnwidth]{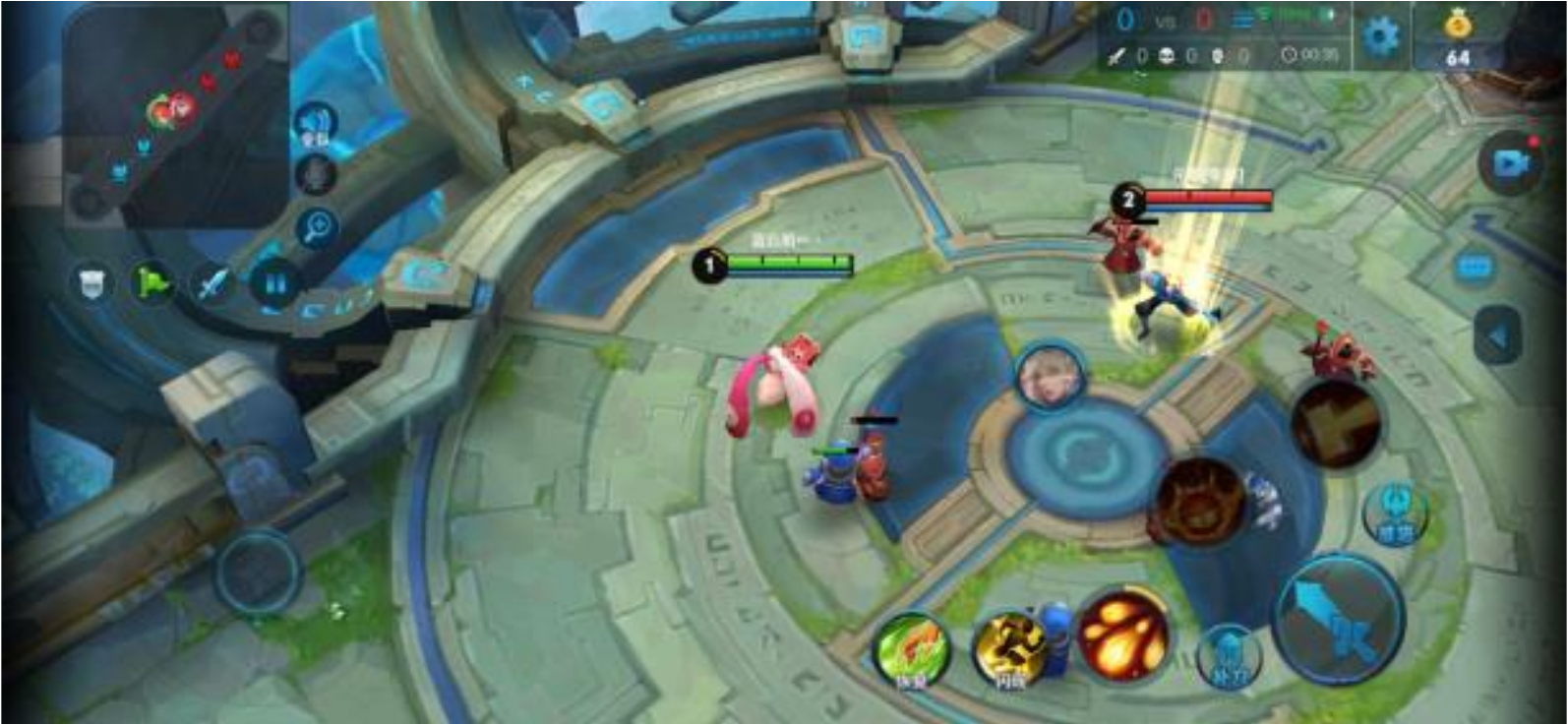}}
\caption{ (a) Screenshot from 5v5 map of {\it KOG}. Players can get the position of allies, towers, enemies in view and know whether jungles alive or not from mini-map. From the screen, players can observe surrounding information including what kind of skills are released and releasing. (b) Screenshot from 1v1 map of {\it KOG}, known as solo mode.}
\label{fig:graph1}
\end{center}
\vskip -0.25in
\end{figure}

MOBA games take up more than 30\% of the online game-plays all over the world, including League of Legends, Dota, {\it King of Glory} ({\it KOG}), and others ~\cite{murphy2015most}. Fig. 1a shows a 5v5 map of {\it KOG}, where players control the motion of heros by controlling the left bottom steer button, while using skills by controling right bottom set of buttons. The upper-left corner shows mini-map, with the blue markers incidating own towers and the red markers incidating the enemies’ towers. Each player can obtain gold and experience by killing enemies, jungling and destroying towers. The ultimate goal of this game is to destroy enemies’ crystal. As shown in Fig. 1b, there are two players in 1v1 map.

	Compared with Atari, the main challenges of MOBA games for us are: (1) \textbf{Game engine or Application Programming Interface (API) is not available for us}. We need to extract features by multi-target detection, and run the game on mobile phones, which is restricated by low computational power. However, the computational complexity can be up to 10$^{20,000}$, while AlphaGo is about 10$^{250}$ ~\protect\cite{OpenAI}. (2) \textbf{Rewards are severely delayed and sparse}. The ultimate goal of the game is to destroy the enemies’ crystal, which means that rewards are seriously delayed. Meanwhile, the rewards are really sparse if we set the rewards of \begin{math}-1/1\end{math} according to the final result loss/win. (3) \textbf{Multi-agents' communication and cooperation are challenging}. Communication and cooperation are crucially important for RTS games especially in 5v5 mode.

	To the best of our knowledge, this paper is the first attempt to propose reinforcement learning in MOBA game, which does not obtain information from API, but captures information from game video directly. We cope with the curse of computational complexity through imitation learning of the macro strategies, and it is hard to train for agents because the rewards in this part are severely delayed and sparse. Meanwhile, we develop a distributed platform for sampling to accelerate the training process and combine the A-Star path planning algorithm to do navigation. To test the performance of our method, we take the game {\it KOG}, a popular mobile MOBA game, as our experiment environment, and systematic experiments have been performed.

	The main contributions of this work include: (1) proposing a novel hierarchical reinforcement learning framework for a kind of mobile MOBA game {\it KOG}, which combines imitation learning and reinforcement learning. Imitation learning according to humans' experience is responsible for macro strategies such as where to go to, when to offend and defend, while reinforcement learning is in charge of micromanipulations such as which skill to release and how to move in battle; (2) developing a simple self-learning method which learns to compete with agent’s past good decisions and come up with an optimal policy to accelerate the training process; (3) developing a multi-target detection method to extract global features composing the state of reinforcement learning; (4) designing a dense reward function and using real-time data to communicate with each other. Experiments show that our agents learn better policy than other reinforcement learning methods.

\section{Related Work}

\subsection{RTS Games}
There has been a history of studies on RTS games such as StarCraft ~\cite{ontanon2013survey} and Dota ~\cite{OpenAI}. One practical way using rule-based method by bot SAIDA achieved championship on SSCAIT recently. Based on the experience of the game, rule-based bots can only choose predefined action and policy at the beginning of a game, which is insufficient to deal with large and real time state space throughout the game, and it is difficult to keep learning and evolving. Dota2 AI created by OpenAI, named OpenAI Five, has made great success by using Proximal Policy Optimization (PPO) algorithm together with well-designed rewards. However, OpenAI Five has used huge computing resources due to lacking of macro strategy. 
	
	Related work has also been done in macro strategies by Tencent AI Lab in game {\it KOG} ~\cite{wu2018hierarchical}, and their 5-AI team achieved 48\% win rate against human player teams which are ranked top 1\% in the player ranking system. However, 5-AI team used supervised learning and the training data can be obtained from game replays processed by game engine and API, which run on server. This method is not possible for us because we don’t have access to the game engine or API, and we need to run on mobile phones.

\subsection{Hierarchical Reinforcement Learning}

Traditional reinforcement learning methods such as Q-learning or Deep Q Network (DQN) is difficult to manage due to large state space in environment, Hierarchical reinforcement learning ~\cite{barto2003recent} tackles this kind of problem by decomposing a high dimensional target into several sub-target which is easier to cope with.

	Hierarchical reinforcement learning has been explored in different scenarios. As for games, somewhat related to our hierarchical architecture exists ~\cite{sun2018tstarbots}, which takes advantage of prior knowledge of a game to design macro strategies, and there is neither imitation learning nor experinenced experts' guidance. There have been many novel hierarchical reinforcement learning algorithms proposed recently. One approach of combining meta-learning with a hierarchical learning is Meta Learning Shared Hierarchies (MLSH) ~\cite{frans2017meta}, which is mainly used in multi-task learning and transfer learning. Hierarchically Guided Imitation Learning/Reinforcement Learning haved been showed effective in speeding up learning ~\cite{le2018hierarchical}, but it needs high-level expert guidence in micro strategies which is hard to design for RTS games.

\begin{figure*}[tb]
\vskip 0in
\begin{center}
\centerline
{\includegraphics[width=1.8\columnwidth]{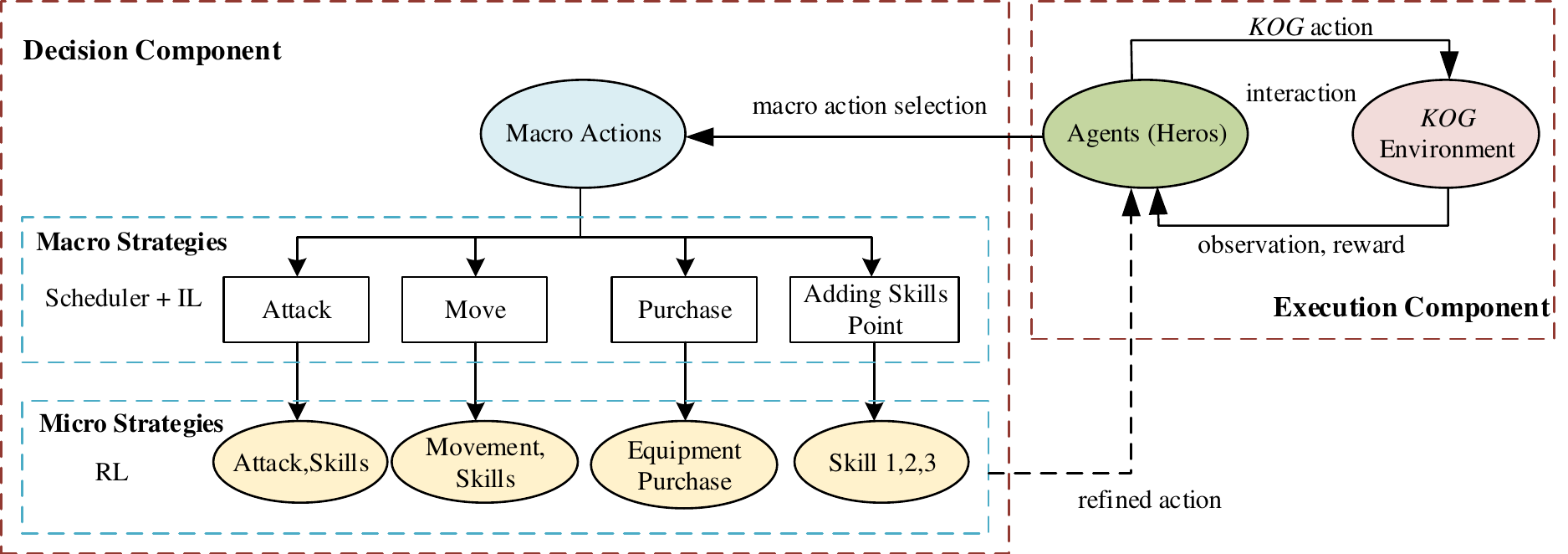}}
\caption{Hierarchical architecture}
\label{figure1}
\end{center}
\vskip -0.25in
\end{figure*}

\subsection{Multi-agent Reinforcement Learning in Games}

Multi-agent reinforcement learning(MARL) has certain advantages over single agent learning. Different agents can complete tasks faster and better through knowledge sharing, and there are some challenges as well. For example, the computational complexity increases due to larger state space and action space compared with single agent learning. Because of the above challenges, MARL mainly focuses on stability and adaption.

	Simple applications of reinforcement learning to MARL is limited, such as no communication and cooperation among agents ~\cite{sukhbaatar2016learning}, lacking of global rewards ~\cite{rashid2018qmix}, and failure to consider enemies’ strategies when learning policy. Some recent studies relevant to this challenge have been investigated. ~\cite{foerster2017stabilising} introduced a concentrated criticism of the cooperative settings with shared rewards. The approach interprets experiences in the replay memory as off-environment data and marginalize the action of a single agent while keeping others unchanged. These methods enable successful combination of experience replay with multi-agent. Similarly, ~\cite{jiang2018learning} proposed an attentional communication model based on actor-critic algorithm for MARL, which learns to communicate and share information when making decision. Therefore, this approach can be a complement for the proposed research. Parameter sharing multi-agent gradient descent Sarsa (PS-MASGDS) algorithm ~\cite{shao2018starcraft} used a neural network to estimate the value function and proposed a reward function to balance the units move and attack in the game of StarCraft, which can be learned from. However, these methods require a lot of computing resources.

\section{Methods}
This section introduces hierarchical architecture, state representation and action definition. Then the network architecture and training algorithm are presented. The reward function design and self-learning method are discussed at last.

\subsection{Hierarchical Architecture}
The hierarchical architecture is shown in Fig. 2. There are four types of macro actions including attack, move, purchase and adding skill points, which are selected by imitation learning. Then reinforcement learning algorithm chooses specific action \begin{math}a\end{math} according to policy $\pi$ for making micro strategies in state \begin{math}s\end{math}. The encoded action is performed and the agent can get reward \begin{math}r\end{math} and next observation \begin{math}s\end{math}$^{'}$ from {\it KOG} environment. \begin{math}R_\pi=\sum_{t=0}^T\gamma^tr_t\end{math} is defined as the discounted return, where $\gamma\in$[0,1] is a discount factor. The aim of agents is to learn a policy that maximizes the expected discounted returns, defined as:
\begin{align}
    J=E_\pi[R_\pi]
\end{align}%

	The Scheduler module designed by observation from game video information is responsible for switching between reinforcement and imitation learning. It is also possible to replace the imitation learning part with high-level expert system for the fact that the data in imitation learning model is produced by high-level expert guidance.

\subsection{State Representation and Action Definition}

\begin{table}
\centering
\begin{tabular}{llllll}
\hline
States  &Dimensionality  &Type \\
\hline
Extracted features     & 116   &  R \\
Mini-map information   & 64$\times$64$\times$3  & R \\    
Current view information     & 84$\times$42$\times$3       & R     \\
Action\_A                &7             & one-hot \\
Action\_M                &9             & one-hot \\
\hline
\end{tabular}
\caption{The dimension and data type of our states and action}
\label{tab:plain1}
\end{table}

\subsubsection{State Representation}

It is an open problem on how to represent the state of RTS games optimally. This paper construct a state representation as inputs to neural network from features extracted by multi-target detection, mini-map information of the game, and current view of the agent, which have different dimensions and data types, as illustrated in Table 1. Current view information is RGB image in the view of the agent, and mini-map information is from RGB image in the upper-left corner of the screenshot.

Extracted features include the position of all heroes, towers, and soldiers, blood volume, gold that the player have and skills released by heroes in the current view, as shown in Fig. 3. All the extracted features are embedded to a 116-dimensional tensor. The inputs at current step are composed of current state information, the last step information, and the last action which has been shown to be useful for the learning process in reinforcement learning. States with real value are normalized to [0, 1].
\subsubsection{Action Definition}
In this game, we define the action into two parts including Action\_M and Action\_A. The motion movement Action\_M includes Up, Down, Left, Right, Lower-right, Lower-left, Upper-right, Upper-left, and Stay still. When the selected action is attack Action\_A, it can be Stay still, Skill-1, Skill-2, Skill-3, Attack, and summoned skills including Flash and Restore. Meanwhile, it is our first choice to attack the weakest enemy when action attack is available for each agent.

\begin{figure*}[tb]
\vskip 0in
\begin{center}
\centerline
{\includegraphics[width=1.8\columnwidth]{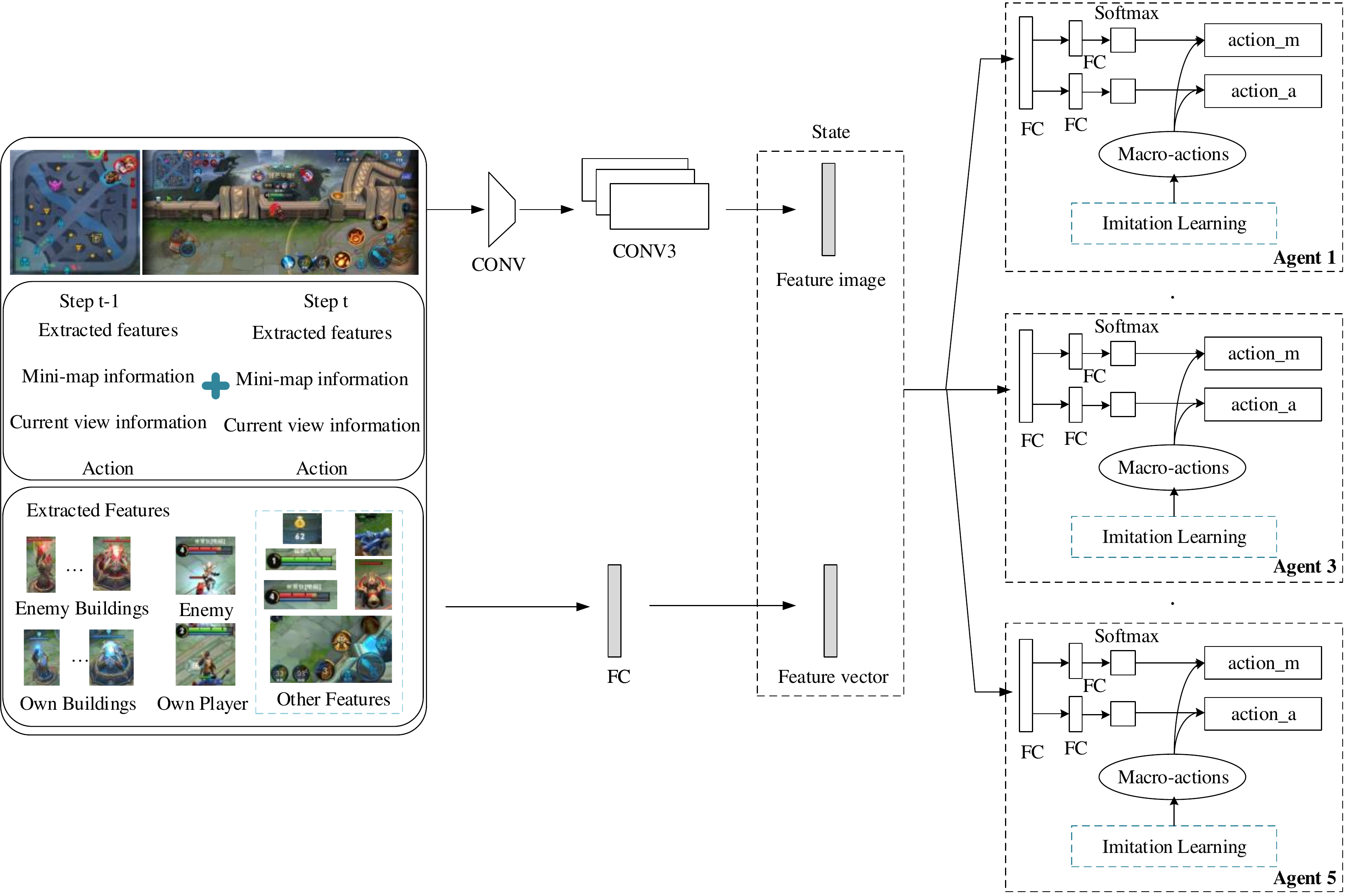}}
\caption{Network architecture of hierarchical reinforcement learning model}
\label{figure2}
\end{center}
\vskip -0.25in
\end{figure*}

\subsection{Network Architecture and Training Algorithm}

\subsubsection{Network Architecture}

Table reinforcement learning such as Q-learning has limitations in large state space situations. To tackle this problem, the micro level algorithm design is similar to proximal policy optimization (PPO) algorithm ~\cite{schulman2017proximal}. Inputs of convolutional network are current view and mini-map information with a shape of 84$\times$42$\times$3 and 64$\times$64$\times$3 respectively. Meanwhile, the extracted features consist of 116-dimensional tensors. We use the rectified linear unit (ReLU) activation function in the hidden layer. The output layer’s activation function is softmax function, which outputs the probability of each action. Our model in game {\it KOG}, including inputs and architecture of the network, and output of actions, is depicted in Fig. 3.

\begin{algorithm}[tb]
\caption{Hierarchical RL Training Algorithm}
\label{alg:algorithm}
\textbf{Input}: Reward function \begin{math}R_n\end{math}, max episodes M, function IL(s) indicates imitation learning model.\\
\textbf{Output}: Hierarchical reinforcement learning neural network.
\begin{algorithmic}[1] 
\STATE Initialize controller policy $\pi$, global state \begin{math}s_g\end{math} shared among our agents;
\FOR {\begin{math}{\it episode} = 1,2,\cdots,{\it M}\end{math}}
\STATE Initialize \begin{math}s_t\end{math}, \begin{math}a_t^m\end{math}, \begin{math}a_t^a\end{math};
\REPEAT 
\STATE Take action [\begin{math}a_t^m\end{math}, \begin{math}a_t^a\end{math}], receive reward \begin{math}r_{t+1}\end{math}, next state \begin{math}s_{t+1}\end{math}, where \begin{math}a_t^m\end{math} indicates a motion movement, and \begin{math}a_t^a\end{math} indicates a motion attack;
\STATE Choose macro action \begin{math}\bm{A_{t+1}}\end{math}from \begin{math}s_{t+1}\end{math} according to \begin{math}IL(s=s_{t+1})\end{math};
\STATE Choose micro action [\begin{math}a_{t+1}^m\end{math}, \begin{math}a_{t+1}^a\end{math}] from \begin{math}\bm{A_{t+1}}\end{math} according to the output of RL in state \begin{math}s_{t+1}\end{math};
\IF {\begin{math}a_{t+1}^i\end{math} $\notin$ \begin{math}\bm{A_{t+1}}\end{math}, where \begin{math}{\it i}=0,\cdots,16\end{math}}
\STATE \begin{math}P(a_{t+1}^i| s_{t+1})=0\end{math};
\ELSE
\STATE \begin{math}\resizebox{.91\linewidth}{!}{$
    \displaystyle P(a_{t+1}^i| s_{t+1})=P(a_{t+1}^i| s_{t+1})/\sum_j P(a_{t+1}^j| s_{t+1}) $}\end{math};
\ENDIF
\STATE Collect samples (\begin{math}s_t, a_t^m, a_t^a, r_{t+1} \end{math});
\STATE Update policy parameter $\theta$ to maximize the expected returns;
\UNTIL {\begin{math}s_t\end{math} is terminal}
\ENDFOR
\end{algorithmic}
\end{algorithm}

\subsubsection{Training Algorithm}
This paper proposes a Hierarchical Reinforcement Learning (HRL) algorithm for multi-agent learning, and the training process is presented in Algorithm 1. Firstly, we initialize our controller policy and global state. Then each agent takes a move and attack action pair [\begin{math}a_t^m\end{math}, \begin{math}a_t^a\end{math}] and receive reward \begin{math}r_{t+1}\end{math} and next state \begin{math}s_{t+1}\end{math}. The agent can obtain both macro action through imitation learning and micro action from reinforcement learning from state \begin{math}s_{t+1}\end{math}. The action probability likelihood is normalized to choose action [\begin{math}a_{t+1}^m\end{math}, \begin{math}a_{t+1}^a\end{math}] from macro action \begin{math}\bm{A_{t+1}}\end{math}. At the end of each iteration, we use the experience replay samples to update parameters of the policy network.

	We take the loss of entropy and self-learning into account to encourage exploration in order to balance the trade-off between exploration and exploitation. Loss formula is defined as:
\begin{align}
\resizebox{.91\linewidth}{!}{$
    \displaystyle
   L_t^M(\bm{\theta})=E_t[w_1L_t^v(\bm{\theta})+w_2N_t^M(\pi,a_t)+L_t^{Mp}(\bm{\theta})+w_3S_t^M(\pi,a_t)]
$}
\end{align}%
\begin{align}
\resizebox{.91\linewidth}{!}{$
    \displaystyle
    L_t^A(\bm{\theta})=E_t[w_1L_t^v(\bm{\theta})+w_2N_t^A(\pi,a_t)+L_t^{Ap}(\bm{\theta})+w_3S_t^A(\pi,a_t)]
$}
\end{align}%
\begin{align}
    L_t(\bm{\theta})=L_t^M(\bm{\theta})+L_t^A(\bm{\theta})
\end{align}%
where \begin{math} L_t^M(\bm{\theta})\end{math} is the loss of action move, \begin{math} L_t^A(\bm{\theta})\end{math} is the loss of action attack. \begin{math}w_1\end{math}, \begin{math}w_2\end{math}, \begin{math}w_3\end{math} are the weights of value loss, entropy loss and self-learning loss that we need to tune, \begin{math}N_t^M\end{math} denotes the entropy loss of action move, \begin{math}N_t^A\end{math} denotes the entropy loss of action attack , \begin{math}S_t^M\end{math} means the self-learning loss of action move, and \begin{math}S_t^A\end{math} means the self-learning loss of action attack. Total loss \begin{math} L_t(\bm{\theta})\end{math} is composed of the loss of move and attack for simply computation.

Value loss \begin{math}L_t^v(\bm{\theta})\end{math}, policy loss \begin{math}L_t^{Mp}(\bm{\theta})\end{math}, and policy loss \begin{math}L_t^{Ap}(\bm{\theta})\end{math} are defined as follows:
\begin{align}
    L_t^v(\bm{\theta})=E_t[(r(s_t,a_t)+V_t(s_t)-V_t(s_{t+1}))^2]
\end{align}%
\begin{align}
\resizebox{.91\linewidth}{!}{$
    \displaystyle
    L_t^{Mp}(\bm{\theta})=E_t[min(r_t(\bm{\theta})D_t^M,clip(r_t(\bm{\theta}),1-\varepsilon,1+\varepsilon)D_t^M)]
$}
\end{align}%
\begin{align}
\resizebox{.91\linewidth}{!}{$
    \displaystyle
    L_t^{Ap}(\bm{\theta})=E_t[min(r_t(\bm{\theta})D_t^A,clip(r_t(\bm{\theta}),1-\varepsilon,1+\varepsilon)D_t^A)]
$}
\end{align}%
where \begin{math}r_t(\bm{\theta})=\pi_{\bm{\theta}}(a_t|s_t)/\pi_{\bm{\theta}_{old}}(a_t|s_t)\end{math}, \begin{math}D_t^M\end{math} and \begin{math}D_t^A\end{math} are advantage of action move and action attack computed by the difference between return and value estimation.

\subsection{Reward Design and Self-learning}
\subsubsection{Reward Design}
Reward function plays a significant role in reinforcement learning, and good learning results of an agent are mainly depending on diverse rewards. The ultimate goal of the game is to destroy the enemies’ crystal. If our reward is only based on the final result, it will be extremely sparse, and the seriously delayed reward leads to slow learning speed. Dense reward gives quick positive or negative feedback to the agent, and can help the agents to learn faster and better. Damage amount of an agent is not available for us since we don’t have game engine or API. In our experiment, all agents can receive two parts of rewards including self-reward and global-reward. Self-reward consists of gold and Health Points (HP) loss/gain of the agent, while global-reward includes tower loss and death of allies and enemies.
\begin{align}
\begin{split}
r_t=&\rho_1 \times r_{self}+\rho_2 \times r_{global}\\
=&\rho_1((gold_t-gold_{t-1})f_m+(HP_t-HP_{t-1})f_H)\\
&+\rho_2(tower_{loss_t}\times f_t+player_{death_t}\times f_d)
\end{split}
\end{align}%
where \begin{math}tower_{loss_t}\end{math} is positive when enemies’ tower is destroyed, and is negative when own tower is destroyed, the same as \begin{math}player_{death_t}\end{math}, \begin{math}f_m\end{math} is a coefficient of gold loss, the same as \begin{math}f_H\end{math}, \begin{math}f_t\end{math} and \begin{math}f_d\end{math}, \begin{math}\rho_1\end{math} is the weight of self-reward and \begin{math}\rho_2\end{math} means the weight of global-reward. The reward function is effective for training, and the results are shown in the experiment section.
\subsubsection{Self-learning}
There are many kinds of self-learning methods for reinforcement learning such as Self-Imitation Learning (SIL) ~\cite{oh2018self} and Episodic Memory Deep Q-Networks (EMDQN) ~\cite{lin2018episodic}. SIL is applicable to actor-critic architecture, while EMDQN combines episodic memory with DQN. However, considering better sample efficiency and easier-to-tune of the system, the proposed method migrates EMDQN to reinforcement learning algorithm PPO ~\cite{schulman2017proximal}. Loss of self-learning part can be defined as follows:
\begin{align}
\begin{split}
    S_t(\pi,a_t)=&E_t[(V_{t+1}-V_H)^2]\\
&+E_t[min(r_t(\theta)A_{H_t},clip(r_t(\theta),1-\varepsilon,1+\varepsilon)A_{H_t})]
\end{split}
\end{align}%
where the memory target \begin{math}V_H\end{math} is the best value from memory buffer, and \begin{math}A_{H_t }\end{math} means the best advantage from it.
\begin{equation}
\resizebox{.91\linewidth}{!}{$
    \displaystyle
V_H=\left\{
\begin{array}{rcl}
&max((max(R_i(s_t,a_t))),R(s_t,a_t)), & {if (s_t,a_t)\in memory}\\
&R(s_t,a_t),                          & {otherwise}
\end{array} 
\right.
$}
\end{equation}%
\begin{align}
    A_{H_t}=V_H-V_{t+1}(s_{t+1})
\end{align}%
where $\it {i}$ $\in$ [1,2,$\cdots$,$\it{E}$], $\it{E}$ represents the number of episodes in memory buffer that the agent has experienced.

\section{Experiments}
The experiment setup will be introduced first. Then we evaluate the performance of our algorithms on two environments: (i) 1v1 map including entry-level, easy-level and medium-level built-in AI, and (ii) a challenging 5v5 map. We analyze the average rewards and win rates during training.

\subsection{Setup}

The experiment setup includes experiment platform and GPU cluster training platform. In order to increase the diversity and quantity of samples, we use 10 mobile phones for an agent to collect the distributed data. Meanwhile, we need to maintain the consistency of all the distributed mobile phones when training. We transmit the collected sample of all agents to the server and do a centralized training, and share the parameters of network among all agents. Each agent executes its policy based on its own states. As for the features obtained by multi-target detection, its accuracy and category are depicted in Table 2, which is adequate for our learning process. Moreover, the speed of taking an action is about 150 Actions Per Minute (APM), comparable to 180 APM of high level player. A-star path planning algorithm is applied when going to someplace. Parameters of \begin{math}w_1\end{math}, \begin{math}w_2\end{math}, and \begin{math}w_3\end{math} are set to 0.5, -0.01, and 0.1 respectively based on preliminary results. The training time is about seven days for agents on one Tesla P40 GPU.

\begin{table}
\centering
\begin{tabular}{llll}
\hline
Category  & Training Set & Testing Set  & Precision\\
\hline
Own Soldier       & 2677  & 382    & 0.6158 \\
Enemy Solider       & 2433  & 380   &  0.6540 \\
Own Tower    & 485  & 79  & 0.9062     \\
Enemy Tower   & 442  & 76    & 0.9091 \\
Own Crystal   & 95  & 17  & 0.9902     \\
Enemy Crystal  & 152   &32   &0.8425   \\
\hline
\end{tabular}
\caption{The accuracy of multi-target detection}
\label{tab:plain3}
\end{table}

\begin{table}
\centering
\begin{tabular}{lllll}
\hline
Scenarios  & AI. 1 & AI. 2  & AI. 3 & AI. 4\\
\hline
1v1 mode      & 80\%  &/     & 52\%    & 58\%\\
5v5 mode    & 82\%  & 68\%   & 66\%   & 60\% \\
\hline
\end{tabular}
\caption{Win rates playing against AI. 1: AI without macro strategy, AI. 2: without multi-agent, AI. 3: without global reward and AI. 4: without self-learning method}
\label{tab:plain4}
\end{table}

\subsection{1v1 mode of game \it {KOG}}

As shown in Fig. 1b, there are one agent and one enemy player in 1v1 map. We need to destroy the enemies' tower first and then destroy the crystal to get the final victory. We draw the win rates and average rewards when agent fights with different level of built-in AI.

\subsubsection{Win Rates}

The results when our AI plays against AI without macro-strategy, without multi-agent, without global reward and without self-learning method are listed in Table 3. 50 games are played against AI. 1, AI. 3 and AI. 4, and the win rates are 80\%, 52\% and 58\% respectively. We have tested about 200 games for each level of built-in AI, and listed the win rates for algorithm PPO and HRL in 1v1 mode against different level of built-in AI, as shown in Table 4. 

\subsubsection{Average Rewards}

Generally speaking, the target of our agent is to defeat the enemies as soon as possible. Fig. 4 illustrates the average rewards of our agent Angela in 1v1 mode when combatting with different enemies. In the beginning, the rewards are low because the agent is still a beginner and doesn't have enough learning experience. However, our agent is learning gradually and being more and more experienced. When the training episodes of our agent reach about 100, the rewards in each step become positive overall and our agent starts to have some advantages in battle. There are also some decreases in rewards when facing high level built-in AI because of the fact that the agent is unable to defeat the Warrior at first. To sum up, the average rewards are increasing obviously, and stay smooth after about 600 episodes.

\begin{figure}[tb]
\vskip 0in
\begin{center}
\centerline
{\includegraphics[width=\columnwidth]{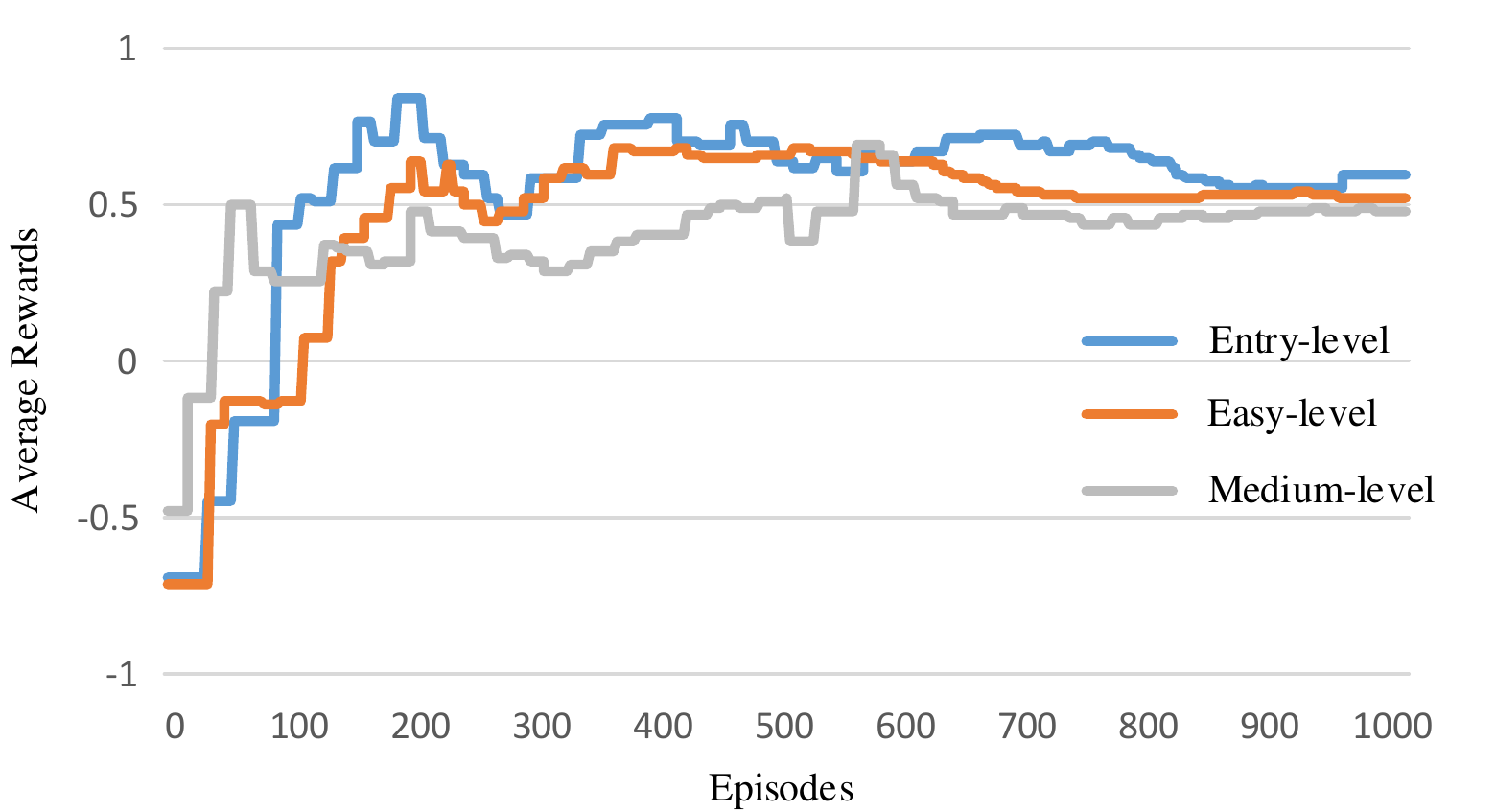}}
\caption{The average rewards of our agent in 1v1 mode during training.}
\label{figure3}
\end{center}
\vskip -0.25in
\end{figure}

\subsection{5v5 mode of game \it {KOG}}
As shown in Fig. 1a, there are five agents and five enemy players in 5v5 map. What we need to do actually is to destroy the enemies’ crystal. In this scenario, we train our agents with built-in AI, and each agent holds one model. In order to analyze the results during training, we illustrate the win rates in Fig. 5.
\subsubsection{Win Rates}
We have plotted the win rates in Fig. 5. there are three different levels of built-in AI that our agents combat with. When fighting with bronze-level built-in AI, agents learn fast and the win rates reach 100\%. When training with gold-level built-in AI, the learning process is slow and agents can’t win until 100 episodes. In this mode, the win rates are about 40\% in the end. This is likely due to the fact that our agents can hardly obtain dense global rewards when playing against high level AI, which leads to hard cooperation in team battle. One way using supervised learning method from Tencent AI Lab obtains 100\% win rate ~\protect\cite{wu2018hierarchical}. However, the method used about 300 thousand game replays with the advantage of API. Another way is using PPO algorithm without macro strategy, which achieves about 22\% win rate when combatting with gold-level built-in AI. Meanwhile, the results of our AI playing against AI without macro strategy, without multi-agent, without global reward and without self-learning method are listed in Table 3. These indicate the importance of each method in our hierarchical reinforcement learning algorithm.

\begin{table}
\centering
\begin{tabular}{llll}
\hline
Algorithm  & Entry-level & Easy-level  & Medium-level\\
\hline
PPO     & 62\%  & 60\%    & 55\%\\
HRL    & 89\%  & 83\%     & 80\% \\
\hline
\end{tabular}
\caption{Win rates for HRL and PPO in 1v1 mode against different level of built-in AI.}
\label{tab:plain4}
\end{table}

\begin{figure}[tb]
\vskip 0in
\begin{center}
\centerline
{\includegraphics[width=\columnwidth]{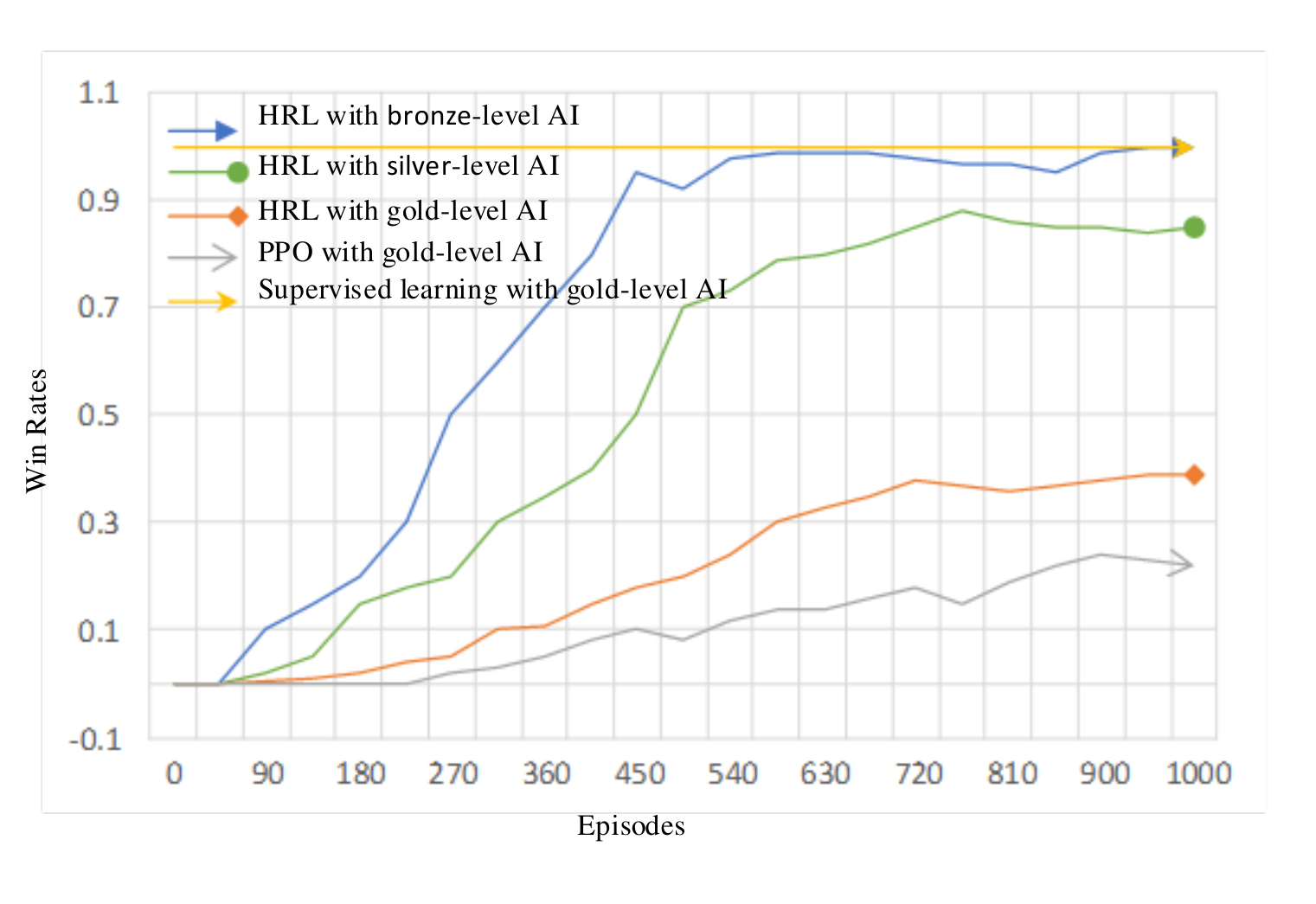}}
\caption{Win rates of our agents in 5v5 mode against different level of built-in AI.}
\label{figure6}
\end{center}
\vskip -0.25in
\end{figure}

\section{Conclusion}
This paper proposed a novel hierarchical reinforcement learning framework for multi-agent MOBA game {\it KOG}, which learns macro strategies through imitation learning and micro actions by reinforcement learning. In order to obtain better sample efficiency, we presented a simple self-learning method, and extracted global features as a part of state input by multi-target detection. We performed systematic experiments both in 1v1 mode and 5v5 mode, and compared our method with PPO algorithm. Our results showed that this hierarchical reinforcement learning framework is encouraging for the MOBA game.

	In the future, we will explore how to combine graph network with our method for multi-agent collaboration

\section*{Acknowledgments}
We would like to thank two anonymous reviewers for their insightful comments and our colleagues, particularly Dr. Yang Wang, Dr. Hao Wang, Jingwei Zhao, and Guozhi Wang, for extensive discussion and suggestion. We are also very grateful for the support from vivo AI Lab.

\newpage
\bibliographystyle{named}
\bibliography{ijcai19}

\begin{thebibliography}{}

\bibitem[\protect\citeauthoryear{Barto and Mahadevan}{2003}]{barto2003recent}
Andrew~G Barto and Sridhar Mahadevan.
\newblock Recent advances in hierarchical reinforcement learning.
\newblock {\em Discrete event dynamic systems}, 13(1-2):41--77, 2003.

\bibitem[\protect\citeauthoryear{Foerster \bgroup \em et al.\egroup
  }{2017}]{foerster2017stabilising}
Jakob Foerster, Nantas Nardelli, Gregory Farquhar, Triantafyllos Afouras,
  Philip~HS Torr, Pushmeet Kohli, and Shimon Whiteson.
\newblock Stabilising experience replay for deep multi-agent reinforcement
  learning.
\newblock {\em arXiv preprint arXiv:1702.08887}, 2017.

\bibitem[\protect\citeauthoryear{Frans \bgroup \em et al.\egroup
  }{2017}]{frans2017meta}
Kevin Frans, Jonathan Ho, Xi~Chen, Pieter Abbeel, and John Schulman.
\newblock Meta learning shared hierarchies.
\newblock {\em arXiv preprint arXiv:1710.09767}, 2017.

\bibitem[\protect\citeauthoryear{Jiang and Lu}{2018}]{jiang2018learning}
Jiechuan Jiang and Zongqing Lu.
\newblock Learning attentional communication for multi-agent cooperation.
\newblock {\em arXiv preprint arXiv:1805.07733}, 2018.

\bibitem[\protect\citeauthoryear{Lin \bgroup \em et al.\egroup
  }{2018}]{lin2018episodic}
Zichuan Lin, Tianqi Zhao, Guangwen Yang, and Lintao Zhang.
\newblock Episodic memory deep q-networks.
\newblock {\em arXiv preprint arXiv:1805.07603}, 2018.

\bibitem[\protect\citeauthoryear{Mnih \bgroup \em et al.\egroup
  }{2015}]{mnih2015human}
Volodymyr Mnih, Koray Kavukcuoglu, David Silver, Andrei~A Rusu, Joel Veness,
  Marc~G Bellemare, Alex Graves, Martin Riedmiller, Andreas~K Fidjeland, Georg
  Ostrovski, et~al.
\newblock Human-level control through deep reinforcement learning.
\newblock {\em Nature}, 518(7540):529, 2015.

\bibitem[\protect\citeauthoryear{Murphy}{2015}]{murphy2015most}
M~Murphy.
\newblock Most played games: November 2015--fallout 4 and black ops iii arise
  while starcraft ii shines, 2015.

\bibitem[\protect\citeauthoryear{Oh \bgroup \em et al.\egroup
  }{2018}]{oh2018self}
Junhyuk Oh, Yijie Guo, Satinder Singh, and Honglak Lee.
\newblock Self-imitation learning.
\newblock {\em arXiv preprint arXiv:1806.05635}, 2018.

\bibitem[\protect\citeauthoryear{Ontan{\'o}n \bgroup \em et al.\egroup
  }{2013}]{ontanon2013survey}
Santiago Ontan{\'o}n, Gabriel Synnaeve, Alberto Uriarte, Florian Richoux, David
  Churchill, and Mike Preuss.
\newblock A survey of real-time strategy game ai research and competition in
  starcraft.
\newblock {\em IEEE Transactions on Computational Intelligence and AI in
  games}, 5(4):293--311, 2013.

\bibitem[\protect\citeauthoryear{OpenAI}{2018}]{OpenAI}
OpenAI.
\newblock Openai five, 2018.
\newblock \url{https: //blog.openai.com/openai-five/}, 2018.

\bibitem[\protect\citeauthoryear{Rashid \bgroup \em et al.\egroup
  }{2018}]{rashid2018qmix}
Tabish Rashid, Mikayel Samvelyan, Christian~Schroeder de~Witt, Gregory
  Farquhar, Jakob Foerster, and Shimon Whiteson.
\newblock Qmix: Monotonic value function factorisation for deep multi-agent
  reinforcement learning.
\newblock {\em arXiv preprint arXiv:1803.11485}, 2018.

\bibitem[\protect\citeauthoryear{Schulman \bgroup \em et al.\egroup
  }{2017}]{schulman2017proximal}
John Schulman, Filip Wolski, Prafulla Dhariwal, Alec Radford, and Oleg Klimov.
\newblock Proximal policy optimization algorithms.
\newblock {\em arXiv preprint arXiv:1707.06347}, 2017.

\bibitem[\protect\citeauthoryear{Shao \bgroup \em et al.\egroup
  }{2018}]{shao2018starcraft}
Kun Shao, Yuanheng Zhu, and Dongbin Zhao.
\newblock Starcraft micromanagement with reinforcement learning and curriculum
  transfer learning.
\newblock {\em IEEE Transactions on Emerging Topics in Computational
  Intelligence}, 2018.

\bibitem[\protect\citeauthoryear{Silver \bgroup \em et al.\egroup
  }{2017}]{silver2017mastering}
David Silver, Julian Schrittwieser, Karen Simonyan, Ioannis Antonoglou, Aja
  Huang, Arthur Guez, Thomas Hubert, Lucas Baker, Matthew Lai, Adrian Bolton,
  et~al.
\newblock Mastering the game of go without human knowledge.
\newblock {\em Nature}, 550(7676):354, 2017.

\bibitem[\protect\citeauthoryear{Sukhbaatar \bgroup \em et al.\egroup
  }{2016}]{sukhbaatar2016learning}
Sainbayar Sukhbaatar, Rob Fergus, et~al.
\newblock Learning multiagent communication with backpropagation.
\newblock In {\em Advances in Neural Information Processing Systems}, pages
  2244--2252, 2016.

\bibitem[\protect\citeauthoryear{Sun \bgroup \em et al.\egroup
  }{2018}]{sun2018tstarbots}
Peng Sun, Xinghai Sun, Lei Han, Jiechao Xiong, Qing Wang, Bo~Li, Yang Zheng,
  Ji~Liu, Yongsheng Liu, Han Liu, et~al.
\newblock Tstarbots: Defeating the cheating level builtin ai in starcraft ii in
  the full game.
\newblock {\em arXiv preprint arXiv:1809.07193}, 2018.

\bibitem[\protect\citeauthoryear{Wu \bgroup \em et al.\egroup
  }{2018}]{wu2018hierarchical}
Bin Wu, Qiang Fu, Jing Liang, Peng Qu, Xiaoqian Li, Liang Wang, Wei Liu, Wei
  Yang, and Yongsheng Liu.
\newblock Hierarchical macro strategy model for moba game ai.
\newblock {\em arXiv preprint arXiv:1812.07887}, 2018.

\end{thebibliography}

\end{document}